# Impact Analysis of Harassment Against Women Using Association Rule Mining Approaches:Bangladesh Prospective

**Bahar Uddin Mahmud[1*], Afsana Sharmin[2]**

[1]*Department of CSE, Feni University, Feni, Bangladesh, Emails: mahmudbaharuddin@gmail.com*
[2]*Department of CSE, CUET, Chittagong, Bangladesh, Email: afsana.cuet@gmail.com*

**Abstract**

In recent years, it has been noticed that women are making progress in every sector of society. Their involvement in every field, such as education, job market, social work, etc., is increasing at a remarkable rate. For the last several years, the government has been trying its level best for the advancement of women in every sector by doing several research work and activities and funding several organizations to motivate women. Although women's involvement in several fields is increasing, the big concern is they are facing several barriers in their advancement, and it is not surprising that sexual harassment is one of them. In Bangladesh, harassment against women, especially students, is a common phenomenon, and it is increasing. In this paper, a survey-based and Apriori algorithm are used to analyze the several impacts of harassment among several age groups. Also, several factors such as frequent impacts of harassment, most vulnerable groups, women mostly facing harassment, the alleged person behind harassment, etc., are analyzed through association rule mining of Apriori algorithm and F.P. Growth algorithm. And then, a comparison of performance between both algorithms has been shown briefly. For this analysis, data have been carefully collected from all ages.

## I. Introduction

These instructions Unwelcome sexual behavior which could be expected to make someone feel offended, ashamed, or intimidated is defined as sexual harassment. It can be physical, verbal, or written. The term sexual harassment came in a 1908 Harper's Bazaar edition where several women who have dealt with sexual harassment wrote their experiences. At that time, many of the published letters discussed several bitter experiences of sexual harassment. Sexual harassment is some sort of behavior that demeans and humiliates an individual based on sex. It is more likely true that Women are the main victims of sexual harassment as they are considered as threats to male status. It is found that women who perform in stereotypically masculine ways (e.g., assertive, dominant, and independent) are more likely to experience harassment.

### I.A. *Definition of Sexual Harassment*

Formally sexual harassment is related to harassment that includes unwelcome or inappropriate rewards in exchange for some sort of sexual favors(M. A. Paludi *et al.,* 1991). Sexual harassment includes a range of actions such as physical conduct, verbal conduct, or nonverbal conduct. The word unwelcome behavior is some sort of critical word. In some circumstances, a victim may agree to participate in a sexual or unwelcome activity, although it is considered to be offensive and questionable. It generally depends on several situations whether the person welcomed a request of sexual activities such as date, sexual comments, jokes, etc. So it completely depends on the victim whether they consider it unwelcome or not(H. B. Philips *et al.,* 1992). Harassment can occur in any place like in the workplace, educational institute, public place, public transport, etc. and women of any ages could be the victim of harassment(K. F. Maria *et al.,*2021)

### I.B. *Situation of Sexual Harassment*

* **Corresponding Author:** Bahar Uddin Mahmud, Lecturer, Department of CSE, Feni University, Feni, Bangladesh; Email: *mahmudbaharuddin@gmail.com*

Sexual harassment can occur in any place such as at own home, educational institutes, public places, public transport, via social media, workplace, etc. And it is often found that the perpetrator has some sort of authority or power on the victim]. The criminal can be of anyone such as family member, friend, co-worker, authority from schools and colleges, casual romantic partner, neighbor, a boss from the workplace, etc. Harassment can occur in various places, and it could occur multiple times depending on several circumstances. And it is found that due to frequent interaction online nowadays, online harassment is increasing at a high rate(B Jahan et al 2022). 2017 PEW research statistics show that on online harassment, 25 percent of women and 13 percent of men between the ages of 18 and 24 have experienced sexual harassment while they are using social media or use another online media for several purposes(M. Duggan, 2020).

### I.C. *Impact of Sexual Harassment*

There are wide ranges of impacts on victims due to sexual harassment. Several kinds of incidents the victim encountered after being sexually harassed, such as anxiety, sleep disturbance, intense fear, depression, ongoing guilt, avoidance behavior, disrupted work-life, headaches, and many more. There is a strong association between workplace sexual harassment and physical harm(C. Harnois *et al.*, 2020). The degree of psychological effect depends on the type of harassment and other circumstances. Both psychological and health effect can occur to a person who faced harassment. Some of the effects that occur are a nightmare, shame, guilt feeling, anger, violence towards the perpetrator, losing confidence, isolation, etc. Even sometimes, the victim is forced to stop his/her regular activity due to harassment.

### I.D. *Sexual Harassment in Bangladesh*

Despite several initiatives and actions, sexual harassment is one of the burning issues in Bangladesh, and one particular gender is the main victim of this activity. Rapid economic growth brings several opportunities for women to expand their potential in many sectors, and women's participation in the job market, social work, and other sector is at an increasing rate. Moreover, the government's several initiatives create consciousness among people about female education, and it shows that the literacy rate of females is notable although less than male. According to the world economic forum Global Gender gap Index, Bangladesh has improved a lot in the scene of discrimination in some fields like the education sector, health sector, etc., but it is a matter of sorrow that the incident of sexual harassment against women is increasing in rapid growth(S. Khatun, 2019). Reports from several organizations show how difficult the current situation is now. About 64 percent of young girls experience sexual harassment in public places. Statistics show that young girls are more vulnerable to sexual harassment. Due to the growth of internet users, reports say young girls are increasingly falling victim to online harassment and abuse(R. Manjoo, 2019). Overall, 84 percent of women are continuously facing harassment in Bangladesh, and it could be anywhere like streets, workplaces, job places, etc. (A. J. Begum, 2018). Women are one of the major employees in the readymade garments sector in Bangladesh. There are several reports of harassment of working women, and it shows how severe the condition is. About 63 percent of garment working women have faced verbal harassment, above 60 percent said about psychological harassment, 24 percent talked about physical harassment, and 11 percent about sexual harassment. Another survey report says about 80 percent of women face sexual harassment at work, and it affects their regular life very badly (J. Pudelek, 2019). Sexual harassment in an educational institute is increasing at an alarming rate, and the nature of the harassment is shockingly severe. About 76 percent of higher education female students have talked about their sexual harassment in post-secondary institutes, and about 45-55 percent of women have faced harassment since the age of 15. School-going students are more vulnerable to sexual harassment, especially female students. And it is shown that the perpetrator behind this incident could be a teacher or any employee from the school.

### I.E. *Purpose of the Work*

There has been lots of research and activity over the few decades to find out sexual harassment-related issues. There are lots of research work that identifies several issues individually, ignoring the overall scenario. Some research work shows health issues related to sexual harassment aging; some talk about another impact. Some research has been done to find individual sector scenarios. In this paper, we tried to cover several facts related to sexual harassment, such as finding the frequent impacts of harassment and finding the age group who are more vulnerable to harassment. We have tried to show a strong association between several impacts of harassment and have shown the association between several parameters via machine learning approaches also have shown the performances of different association rule mining algorithms.

## II. Related Work

Many research works have been done, and many more are continuing to find out several aspects of sexual harassment against women. One of the major concerns of women is some sort of depressive symptoms due to due to several factors (H. Z. Dahlqvist *et al.,* 2020). (Therese Skoog *et al.,* 2019) have discussed several dimensions of sexual harassment such as victimization, perpetration, impacts, risk during transition from childhood to adolescence which is a very vital period of psychological development. In Bangladesh, females aged under 18 are the most vulnerable group that faces sexual harassment frequently. School and colleges going girls are frequently facing sexual harassment, and most of the time, it remains unknown. As a conservative country, most of the time, women do not share their occurrences with others and face the consequences and effects alone, which makes the situation riskier for them. (K. F. Maria *et al.,* 2021) have discussed the degree of depressive symptoms due to sexual harassment in the workplace. It shows the degree of impact is more when harassed by a superior one in the office than by customers. (F. Begum *et al.,* 2020) have discussed several aspects of garments workers' sexual harassment, and more surprisingly, young women are the main target of harassment. (B. Sivertsen *et al.,* 2019) have discussed several types of sexual harassment faced by colleges and university-going female students in Norway, and the most common forms of sexual harassment are comments about your body, unwanted touching, grabbing, hugging, kissing, etc. And it is more likely the same here in Bangladesh, and in some cases, it is a more severe condition here. (M. Z. AHMED, 2020) has shown several forms of sexual harassment in the perspective of private and public universities, and according to statistics, it is found that not only the fellow students but also good numbers of teaching staff are involved in this incident. There is a long list of impacts that have been listed so per because of harassment. Sexual harassment has a great economic impact on working women, and there is a huge psychological impact, too (H. Mclaughlin *et al.,* 2021). (J. M. Wolff *et al.,* 2018) discussed relations of sexual harassment and psychological distress such as anger, depression, etc., and found a strong correlation between several impacts and sexual harassment. In contrast, research by (T. K. Kim et al., 2017) on female military personnel of the Republic of Korea Armed Forces shows that Mental trauma is one of the great consequences of harassment, and the unmarried group is more vulnerable. However, there are some contradictions on various methods of finding consequences of sexual harassment. There are discussions on several consequences of the occurrences like anxiety, lower job/academic performance, absenteeism, drug, and alcohol abuse, and at the most extreme, suicide that is some impact reported. However, it is shown that occurrences of some impacts are closely related to the geographical area of victims also. Several reports say that in Bangladesh, the extreme impact is a suicide attempt or commit suicide of the victims. In the FRA (2020) study, women reported feeling anger, annoyance, embarrassment, shame, and fear. Over the longer term, women experienced feeling vulnerable, anxious, having difficulties in relationships and with sleeping, and, in some cases, being depressed.

## III. Survey and Sampling

A survey is one kind of observation technique that generally apply to a group of people to get an overall idea about their feelings opinion on a particular matter. Survey has several uses in a different fields. Basically, the survey method is used in several fields depending on the requirement. Its goal could be limited or widespread. To analyze the behavior of a particular human being, psychologists often do a survey on a particular group of people. The government does a survey to find several issues related to common people and does a survey to find existing problems and public opinion. It is a great way to have an overall idea of mass reaction. Technology is growing faster, and several technological tools make life easier. The medical health sector uses the survey method in greater numbers than others. Nowadays, physicians, nurses, therapists use several survey techniques to find patterns of clinical problems (M. L.Williams, 1997, J. Han *et al.*, 2001). Sampling is one of the vital parts of a survey, and before doing sampling, several issues must have been taken into consideration. Sampling refers to a group of people among the entire population, and sampling reduces the cost and time of the survey.

### III.A *Survey Design*

Survey design has several methods or techniques based on requirements. There should be a proper survey method and proper planning for sampling. Depending on the requirement, a survey is created with a particular number of predetermined questions. Moreover, The comparison of attitudes of different people, as well as their attitude changes, can be made through the survey. A good sample selection plays a vital role because it represents the overall population. Survey research could be both quantitative and qualitative or sometimes using both techniques Social and psychological field is the most common field where surveys are frequently used because the survey is mostly popular for analyzing human behavior (Singleton and Straits, 2009).

### III.B *Sample Selection*

Sample selection is one of the essential parts of survey research, and both cost and time are related to sampling. There are good numbers of sampling techniques that include probability sampling, nonprobability sampling, etc. Probability sampling again has several types, such as Simple Random Sampling, Systematic Sampling, Stratified Sampling. And also Non, probability sampling methods include some types such as convenience sampling, quota sampling, and purposive sampling. Sampling probably has its own characteristics, such as every element must be defined as a known nonzero probability of being sampled, and there might be some random selection on some points. Nonprobability sampling refers to the selection of a population from a particular area, and there is a huge chance that some element or some area would be completely ignored. All sampling has its individual uses depending on several factors like behavior and quality of the sample, Availability of supporting information, requirements, accuracy required, detail analysis, expenses, and other issues.

### III.C *Data Set*

After the collection of data, it is organized according to research requirements. The data set is separated according to age group and age when first experienced harassment. As both of these data is important to find out the several impacts of sampling of these data is crucial. Overall data is divided into several categories based on ages of facing harassment such as under 18, 18-24, 25-34, 35-44, above 45. Also, other parameters of data, such as details about the perpetrator's location of harassment, take into note to find several sides of the survey. It is important to sample data wisely to get more accurate results. As all ages groups consider

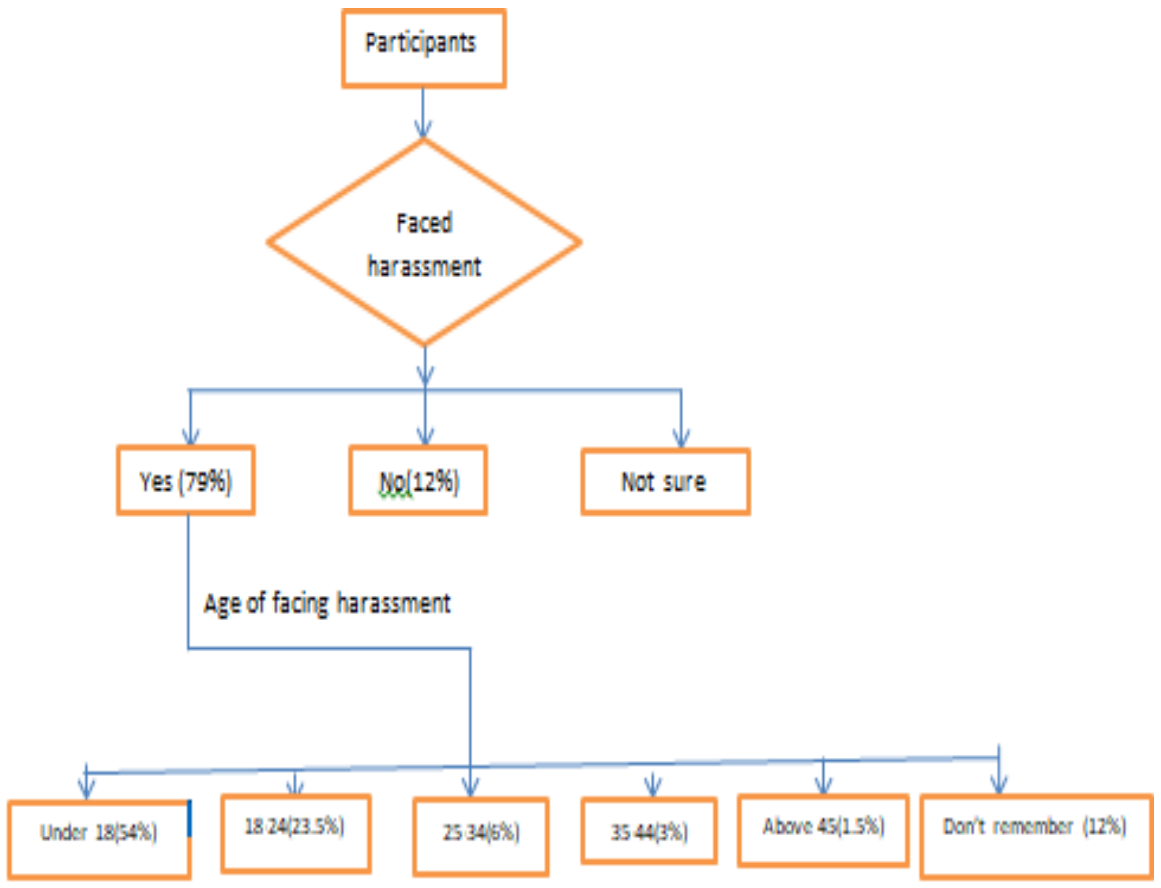

**Figure 1:** Flow diagram of collected data

For the analysis, it is expected that the results will be more accurate. Fig. 1 shows the visual representation of collected data.

### III.D *Data Collection*

Data is collected through a questionnaire both online and offline. In a time of data collection, our main concern was diverse data from several parts of Bangladesh. Diversity of participants was needed for quality research. Our target group was women of different ages. We asked them several questions related to sexual harassment, such as impacts of harassment, who consider both the demography and geography of subjects in a balanced way. We have organized the data age-wise in (Table. 1) as our goal is to analyze impacts by age.

**Table 1.** : Different age groups facing harassment

| Total participants 2100 | |
|---|---|
| Age of facing harassment | Participant distribution |
| >18 | 1169 |
| 18-24 | 577 |
| 25-34 | 180 |
| Above 35 | 154 |
| Don't remember | 220 |

## IV. Association Rule Mining

Initially, association rule mining was one of the popular techniques to analyze the associative pattern of market basket data. This type of analysis helps several big retailers and chain shops to design their product selling strategies based on customers' buying behavior. This technique applies to transactions where transaction consists of a set of items purchased by the customer. After that, association rule mining techniques become popular in many other domains, such as genetic data analysis credit card fraud detection. In every domain, data is analyzed to find the pattern and frequent set of data. There is many association rule mining algorithm in the data mining field. Two of them are Apriori and FP-growth. Apriori follows a breadth-first-search strategy, while FP-growth follows a depth-first search strategy.

## V. Methodology

Following steps have been followed to figure out the expected output.

1. Preprocessed the raw data and converted the raw file into CSV format.
2. Sorting the data as our system requires.
3. Applied Apriori algorithm-based approach to analyze the impact.

is involved in harassment, major location of harassment, their awareness about help center, etc. Before collecting our data, we carefully

4. Applied FP-growth algorithm-based approach to finding frequent itemsets.
5. Made comparison between apriori approach and fp-growth to analyze the performance based on our collected data

V.A. *Apriori algorithm-based classification*

For a big volume of data to find out hidden and targeted patterns, data classification techniques are much needed(G. Shankar *et al.*, 2013). The data mining technique is to find out patterns of data in huge databases, which helps to gain knowledge about the data and make decisions (R. Agrawal *et al.*, 2019). Data mining task could be automatic or semiautomatic analysis and apply to large quantities of data to find out an unknown or interesting group of data records that are known as cluster analysis, abnormal records known as anomaly detection, and frequent data in records known as association rule mining technique of Apriori algorithm is one of the important mining technique in the data mining field. Association rule mining is a powerful technique to find out relationships among several datasets in a large database and extract more frequent data set based on given conditions. It generally correlates one set of items with other sets of items in the database. A minimum support threshold and a minimum confidence threshold find all the rules in the data set that satisfy the specified support and confidence thresholds. Let I be a set of items, D be a data set containing transactions (i.e., sets of items in I), and t be a transaction. An association rule mined from D will be of a form $X \rightarrow Y$, where $X, Y \subset I$ and $X \cap Y = \emptyset$. The support of the rule is the percentage of transactions in D that contain both X and Y. The confidence is, out of all the transactions that contain X, the percentage that contains Y as well. Confidence of a rule can be computed as support $\{X \cup Y\} \div$ Support$\{X\}$. The confidence of a rule measures the strength of the rule (correlation between the antecedent and the consequent), while the support measures the frequency of the antecedent and the consequent together. An iterative approach is known as level-wise search used by apriori, and prior data is analyzed to get later frequent itemsets of data such as K-itemsets are used to find out (K+1)-itemsets. It is obvious that all nonempty subsets of a frequent itemset must also be frequent. There is a two-step process in the Apriori algorithm first one is The Join step, and the second one is The Prune step. Join step denotes that to find further frequent itemset suppose L.K., a set of candidate K-itemsets is generated by joining Lk-1 with it, and it is denoted as Ck. Suppose to find itemsets L3, and we have to consider all L3-1 itemsets and join with L3. Prune step denotes to reduce the size of the set of candidates Ck. Ck is a superset of Lk-1. That means itemsets in Ck may or may not be frequent, but all frequent sets are in there. Now, after getting frequent item sets, we have to generate strong association rules. Those are the strong association rules that satisfy both minimum support and minimum confidence. From minimum support, we will get frequent itemsets. Here is the process how to deal with constraint minimum confidence. For getting association rules, we follow the following steps: All nonempty subsets s of frequent itemsets l must be generated. For every nonempty subset s of l, output s=> (l-s) , if support count(l)/support count(s) >= min conf (1) ,

where minimum confidence is given a threshold value.

V.B. *Impact classification based on age*

We have collected a total of 2100 data from across the country of different ages. The raw data file was then refined through several categories and converted into CSV format to make it compatible with our software. The CSV file is converted in a more comfortable format to run the program. Data is formatted in such a manner that every single response is sorted on a list with the items comma-separated, removing the new lines and making a new list in CSV format. Firstly the apriori algorithm is used to find a frequent list of impacts among several impacts. And then, association rules are generated using minimum support and confidence. Suppose minimum support for every impact is 200 and minimum confidence is 40 PERCENT. Frequent sets have been generated using minimum support and minimum confidence. First, read the CSV file and generate it in python list format. Data has to be sorted as per requirements. Now all elements have been visited and made a count list of all elements and stored in a dictionary. Now the comparison between the support count of all elements with minimum support has been made and store the new data in another dictionary and finally made a set called L1() to avoid repetition.

V.C. *Experimental Work*

Several python libraries were used, such as pandas, NumPy, CSV, etc., for the experiment. Below is the explanation of our algorithmic approach has given :

i. List generation of each individual item C1(): Read the data from the CSV file and make a list of it .then counts of all individual data have been stored in a dictionary.
ii. Generation of 1 frequent itemset - L1(): After listing out all items now, all counts are compared with the minimum

Support and the values had support greater than the support threshold stored in a new dictionary.

iii. Generation of 2 frequent itemsets-C2(): Now, the system automatically traversed through all itemsets of C1 to find two itemsets that are identical.
iv. Generation of L2(): all itemset found in C2 further check if they have existed in an individual itemset, then they are added to list L2 and then thresholded by minimum support. And then, a function was used to find the length of all two frequent itemsets.
v. Generation of L2(C2, data):} C2 <- L2 <- L(length of individual two frequent items The process repeated for three and more frequent itemsets and every time generate

a new dictionary by appending the old dictionary values and comparing with the threshold value

vi. Association rules generation(): now, all frequent items in the final list generated their total support, and all combinations of frequent items have been considered by splitting between the left and right manner and generated support of these combinations. If total support/combination support is greater than the minimum confidence value, these are added to the list of rules, and finally, an output file of rules was generated. Table. 2 and Table. 3 represents the data with total support and data with minimum support, respectively.

**Table 2.** : support count of attributes

| Total participants 2100 | |
|---|---|
| Attribute | Support *count* |
| Anxiety | 1060 |
| Intense fear | 618 |
| Ongoing fears | 860 |
| Ongoing guilt feeling | 168 |
| Depressions | 837 |
| Sleep disturbances or Nightmares | 420 |
| Avoidance behaviors | 84 |
| Headaches | 168 |
| Disrupted work life | 419 |
| Face difficulties with communication | 309 |
| intimacy and enjoyment of social activities | 287 |
| Degradation of performances in study or work | 508 |
| Under 18 | 1169 |
| 18-24 | 577 |
| 25-34 | 180 |
| Above 35 | 154 |
| Don't remember | 220 |

**Table 3.** : minimum support count

| Total participants 2100 | |
|---|---|
| Attribute | Support *count* |
| Anxiety | 1060 |
| Intense fear | 618 |
| Ongoing fears | 860 |
| Depressions | 837 |
| Sleep disturbances or Nightmares | 420 |
| Disrupted work life | 419 |
| Face difficulties with communication | 309 |
| intimacy and enjoyment of social activities | 287 |
| Degradation of performances in study or work | 508 |
| Under 18 | 1169 |
| 18-24 | 577 |
| Don’t remember | 220 |

Now in the second step, go through all itemsets of Lk to find two itemsets that are identical. Then the data has been stored in a list C.K. in a sorted manner and made a set of C.K. to avoid repetition. If itemset in Ck belongs to an individual item list, it has been added to list Ct, and its support is updated by 1 using the minimum support new dictionary of items has been created from the old dictionary. Table. 4 represents the data of two frequent itemsets generated by the apriori algorithm. And Table. 5 represents two frequent data with minimum support.Table.6 and Table. 7 presents the data of three frequent items. And Table 8 represents data with four frequent items.

**Table 4.** : Two frequent attributes

| |
|---|
| (’18-24’, ’Intense fear’) |
| (’Anxiety’, ’Ongoing fears ’) |
| (’18-24’, ’Ongoing fears ’) |
| ('Anxiety,' 'Degradation of performances in study or work') |
| ................. |
| .................. |
| (’18-24’, ’Depressions’) |
| (’Anxiety’, ’Under 18’) |
| (’Anxiety’, ’Sleep disturbances or Nightmares’) |
| (’Anxiety’, ’Disrupted work life’) |
| ................. |
| ................... |
| (’18-24’, ’Anxiety ’) |
| (’Anxiety’, ’Face difficulties with communication’) |
| (’Anxiety’, ’Ongoing fears’) |
| (’Anxiety’, ’I dont remember’) |
| (’Intense fear’, ’Ongoing fears ’) |
| ............... |
| ................. |
| (’Face difficulties with communication’,’ Under 18’) |
| (’ intimacy and enjoyment of social activities’, ’Ongoing fears’) |
| (’Face difficulties with communication’, ’Under 18’) |
| (’Intense fear’, ’Under 18’) |
| ..................... |
| ...................... |
| ('Intense fear', 'Sleep disturbances or Nightmares') |
| ('Intense fear', 'Ongoing fears') |
| ('Anxiety ', 'Intense fear') |
| ('Anxiety ', 'Depressions') |
| ('Face difficulties with communication,' 'I don't remember) |

**Table 5.** : Two frequent attributes with minimum support count

| Total participants 2100 | |
|---|---|
| Attribute | Support_*count* |
| ('Face difficulties with communication,' 'Intimacy and enjoyment of social activities) | 287 |
| ('Depressions,' 'Face difficulties with communication') | 220 |
| ('Anxiety,' 'Sleep disturbances or Nightmares') | 221 |
| ('Depressions', 'Disrupted work life') | 286 |
| ('Intense fear', 'Under 18') | 396 |
| ...................... | .... |
| ...................... | .... |
| ('Anxiety', 'Under 18') | 639 |
| ('Ongoing fears', 'Under 18') | 595 |
| (' intimacy and enjoyment of social activities','Face difficulties with communication') | 287 |
| ('Anxiety', 'Ongoing fears') | 506 |
| ('Anxiety', 'Intense fear') | 418 |
| ('Depressions', 'Under 18') | 595 |
| ('Anxiety', 'Depressions') | 529 |
| ('18-24', 'Anxiety') | 289 |
| ...................... | .... |
| ...................... | .... |

Now traverse through all previous frequent itemsets having two items to find the three items candidate and again check the support count with minimum support and store in another dictionary.

**Table 6.** : Three frequent attribute set

| |
|---|
| ('Anxiety', 'Intense fear', 'Ongoing fears') |
| ('Anxiety', 'Depressions', 'Intense fear') |
| ('Anxiety', 'Intense fear', 'Under 18') |
| ('Anxiety', 'Intense fear', 'Ongoing fears') |
| ('Anxiety', 'Depressions', 'Ongoing fears') |
| ('Anxiety', 'Ongoing fears', 'Under 18') |
| ('Anxiety', 'Depressions', 'Intense fear') |
| ('Anxiety', 'Depressions', 'Ongoing fears') |
| ('Anxiety', 'Depressions', 'Under 18') |
| ('Anxiety', 'Intense fear', 'Under 18') |
| ('Anxiety', 'Ongoing fears', 'Under 18') |
| ('Anxiety', 'Depressions', 'Under 18') |
| ('Degradation of performances in study or work', 'Depressions', 'Under 18') |
| ('Depressions', 'Ongoing fears', 'Under 18') |
| ('Degradation of performances in study or work,' 'Depressions,' 'Ongoing fears') |
| ('Intense fear', 'Ongoing fears', 'Under 18') |
| ('Degradation of performances in study or work,' 'Ongoing fears,' 'Under 18') |

**Table 7.** : Name of the Table that justify the values

| Total participants 2100 | |
|---|---|
| Attribute | suppor tcount |
| ('Anxiety', 'Intense fear', 'Ongoing fears') | 264 |
| ('Anxiety', 'Ongoing fears', 'Under 18') | 220 |
| ('Anxiety', 'Depressions', 'Ongoing fears') | 352 |
| ('Anxiety', 'Intense fear', 'Under 18') | 286 |
| ('Anxiety', 'Depressions', 'Intense fear') | 242 |
| ('Anxiety', 'Depressions', 'Under 18') | 397 |
| ('Anxiety', 'Depressions', 'Under 18') | ..... |
| ('Anxiety', 'Depressions', 'Under 18') | .... |
| ('Degradation of performances in study or work', 'Depressions', 'Under 18') | 243 |
| ('Depressions', 'Ongoing fears', 'Under 18') | 352 |
| ('Intense fear', 'Ongoing fears', 'Under 18') | 220 |

**Table 8.** : Name of the Table that justifies the values

| Total participants 2100 | |
|---|---|
| Attribute | suppor t *count* |
| ('Anxiety', 'Depressions', | 220 |
| ’Disrupted work life’, ’Ongoing fears’) | |
| (’Anxiety’, ’Depressions’, ’Ongoing fears’, ’Under 18’) | 264 |

V.D. *Association Rule*

Total support has been calculated for each itemset in the frequent item list. Then all possible combination of itemsets has been made by splitting them, and support has been generated for this combination from the dictionary. Then it has been added as a rule if the calculated support is greater than minimum confidence and written in a list. Table. 9 shows the association rules between different impacts and age groups.

**Table 9.** : Association rules

| Rules | Confidence | Result Status |
|---|---|---|
| [’Ongoing fears’] -> [’Under 18’] | (0.6918604651162791) | Accepted |
| [Intimacy and enjoyment of social activities’] ->[’Face difficulties with Communication’] | (1.0) | Accepted |
| [’Under 18’] -> [’Depressions’] | (0.3829787234042553) | Rejected |
| … …… … | ……….. | …….. |
| … …… …. | ……….. | ……… |
| [’Intense fear’] -> [’Anxiety’] | (0.6763754045307443) | Accepted |
| [’Anxiety’] -> [’Under 18’] | (0.6028301886792453) | Accepted |
| [’Ongoing fears’] -> [’Anxiety’] | (0.5883720930232558) | Accepted |
| [’Anxiety’,’Depressions’] -> [’Under 18’] | (0.4990566037735849) | Accepted |
| [’Anxiety’,’Intense fear’] -> [’18-24’] | (0.3665432612345678) | Rejected |
| [’Anxiety’,’ongoinging fears’] -> [’18-24’] | (0.2978723404255319) | Rejected |
| [’Intense fear’,Under 18] -> [’Anxiety’] | (0.5843274132543268) | Accepted |
| [’Uder18’]->[’Depressions’] | (0.3829787234042553) | Rejected |
| … | ……… ….. | ……… … |
| … | ……….. | ……….. |
| [’Depressions’, ’Under 18’] -> [’Anxiety’] | (0.6111111111111112) | Accepted |
| [’Anxiety’,’Ongoing fears’] -> [’Depressions’] | (0.5674352345678934) | Accepted |
| … | ……… …. | ……… |
| … | ……….. | ………. |

## VI. FP-Growth Algorithm Based Mining

Unlike the apriori technique, fp-growth finds frequent itemsets without generating candidates. The first frequent-pattern tree, or F.P. tree, has been created compressing the database containing frequent items(L. Zhichun *et al.*, 2008). FP-Growth algorithm maintains the information of associations between itemsets and compresses the database to generate an F.P. tree (C. Jun et al., 2013). The main technique of FP-Growth does not generate candidate itemsets in the process of mining and improves efficiency.

VI.A *FP-Growth Based Rule Mining*

Minimal support for every item is 200, and minimum confidence is 40 percent. First, a list has been created based on the support count of the items in the item list, and then the most important part is to arrange the itemsets in ascending order based on support count.

VI.B *Building F.P. Tree*

To construct the fp tree, null has been taken as a root node. Now every item set has been compared with the last table and arranged in descending order, and now all items are added as the left child of the root node. Now next itemsets have been checked, and if the items are present in the child node of the root, then increment it; otherwise, add it as a new child. This process continued for every item in an itemset and finally for all itemsets in the database.

### VI.C *Conditional Database*

After constructing the fp tree, the conditional database has been created. For conditional database least support count item is considered first. And find the path of traversing from the root node to that node. And all items have been written on a list along with their support count. Now a frequent itemset table is generated for the conditional leaf. And all item less than the minimum support count has been removed, and the rest of the data appended in a list to generate a frequent pattern.

### VI.D *Steps Involved in the Experiment*

We have used several python libraries such as pandas, NumPy, CSV collection for our experimental work. Below is the algorithmic approach that we followed in our experiment to find frequent impact items from our data list.

```
Input: Datasets in CSV format and the threshold
value
Output: Frequent pattern and conditional fp tree
//display the fp tree or conditional fp tree in a
nested list
FUNCTION fp tree list(item,frequency,parent
node,child node linking_pointer)
Show the name and frequency of items.
IF the length of Cn>0, then
show length;
FOR total_Cn:
display the values;
//For any children of the node, call the function
recursively
Call fp tree list();
END FOR;
END IF;
//Writes the frequent itemsets to a CSV file
FUNCTION _to_file(data):
open(output_file_name, "w”);
writer <- csv.writer(f, delimiter=',')
FOR row in data:
writer.writerows([[row]])
END FUNCTION
END FOR
//The most recent node is linked to the previous
node with the same name
FUNCTION same_item_update(same_item,
current_node):
# Traversing
WHILE (same_item.link != None):
same_item <- same_item.link
END WHILE
same_item.link <- current_node
END FUNCTION
```

After that first database scan occurred and scanned to create the frequent item dictionary and deleted the values below a threshold value again; the database scanned the second time and sorted the item according to their frequency; also, if the two-item have the same frequency, we have arranged them alphabetically. And then, it was sent to create the fp tree, and the below algorithm shows the processes, and it works recursively.

```
// function recursively creates the FP-Tree for
each itemset.
FUNCTION fp_tree_creation(initial_node,
itemsets, same_item):
IF child is present
init_node.child.freq++
ELSE create new node for child and add to its Pd
END IF-ELSE
IF similar_item != new node then update table
ELSE
Traverse till the last similar node, and update the
new node
END IF-ELSE
IF length(itemsets) > 1:
call fp_tree_creation(initial_node, itemsets,
same_item)
END FUNCTION
```

Again above algorithm has been running with an extra condition

IF frequent items are similar: no update in the conditional tree.

Now a conditional fp tree has been generated by using the table found from the above algorithm to find frequent items pattern in the fp growth approach. A conditional fp tree has been created for every item in the table, and itemsets below the threshold value have been removed.

```
FUNCTION
overall_frequent_pattern(same_items,threshold)
//pass the table data and threshold value as
argument.
```

FOR key, value in same_items.datasets:
go through every item in the previous table;
WHILE (value!= null)
initialize Cd=value & find frequency;
WHILE Cd.Pd!= None:
then traverse through child node to parent node;
and append the name and value;
END FOR
//After completing the particular value in a link, the next link will be generated, and the whole path is added to the conditioning path. A frequent itemset dictionary is created for the child node.
OUTER LOOP FOR r in condition_base:
INNER LOOP FOR column in row:
IF column[0] not in Cd_frequency:
Cd_Frequency[col[0]] = col[1]
ELSE:
Cd_Frequency [col[0]] = Cd_Frequency [col[0]] +col[1]
END INNER FOR
END OUTER FOR

IF Cd_Frequency< threshold_value:
Remove the item;
END IF
//For every transaction in the condition_base, the items are stored
OUTER LOOP FOR row in condition_base:
generate a list of the frequent item;
INNER LOOP FOR column in a row:
IF col[0] in Cd_Frequency:
stores only the name of the item
stores both name and frequency
END INNER LOOP
END OUTER LOOP
END FUNCTION

Table. 10 represents the frequent pattern of data on the conditional database generated by the F.P. tree. Table. 11 represents the frequent pattern of data to generate strong association rules with a minimum confidence value.

**Table 10.** : Frequent pattern of data

| |
|---|
| "['Intimacy and enjoyment of social activities, 'Face difficulties with communication |
| ['18-24'] |
| ['Anxiety '] |
| ['Degradation of performances in study or work'] |
| ................. |
| .................. |
| "['Intense fear', 'Anxiety', 'Under 18']" |
| "['Intense fear', 'Anxiety']" |
| "['Intense fear', 'Under 18']" |
| "['Ongoing fears', 'Anxiety', 'Under 18']" |
| ................. |
| ................... |
| "['Ongoing fears', 'Anxiety']" |
| ['Ongoing fears'] |
| "['Ongoing fears', 'Anxiety', 'Under 18']" |
| ['Sleep disturbances or Nightmares'] |
| ['Under 18'] |
| ............... |
| ................. |
| "['Intense fear', 'Anxiety', 'Under 18']" |
| (' intimacy and enjoyment of social activities', 'Ongoing fears') |
| "['Intense fear', 'Under 18']" |
| "['Ongoing fears', 'Anxiety']" |
| ..................... |
| ...................... |

**Table 11.** : Association rules from FP-Growth algorithm.

| Rules | Confidence | Status |
|---|---|---|
| ('Anxiety', 'Depressions') ( 357 ) —> Under 18 ( 987 ) | (0.6470588235294 118 ) | Accepted |
| ('Anxiety', 'Under 18') ( 399 ) —> Depressions ( 651 ) | (0.5789473684210 527 ) | Accepted |
| ('Depressions' , 'Under 18') ( 378 ) —> Anxiety ( 735 ) | (0.6111111111111 112 ) | Accepted |
| Anxiety ( 735 ) —> Under 18 ( 987 | (0.5428571428571 428 ) | Accepted |
| Under18(987) ->Anxiety( 735 ) | (0.4042553191489 3614 ) | Accepted |
| Intense fear ( 399 ) —> Anxiety ( 735 ) | (0.5263157894736 842 ) | Accepted |
| Intense fear ( 399 ) —> Under 18 ( 987 ) | (0.6315789473684 21 ) | Accepted |
| Depressions ( 651 ) —> | (0.5806451612903 226 ) | Accepted |
| Under 18 ( 987 | | |
| Anxiety ( 735 ) —> Depressions ( 651 ) | (0.4857142857142857 ) | Accepted |
| Depressions ( 651 ) —> Anxiety ( 735 ) | (0.5483870967741935 ) | Accepted |
| Face difficulties with communication ( 294 ) —> Intimacy and enjoyment of social activities ( 210 | (0.7142857142857143 ) | Accepted |
| Ongoing fears ( 567 ) —>Under 18 ( 987 ) | (0.5185185185185185 ) | Accepted |
| Ongoing fears ( 567 ) —> Anxiety ( 735 ) | (0.4074074074074074 ) | Accepted |
| ………… | ……………… | …….. |

## VII. Research Question

Women of different ages have been considered for this research. However, teenagers are the main victim of several types of harassment. For our experiment, we have asked several questions to women of several ages via an online questionnaire and also in hard format by sharing a questionnaire form with them. We have collected data from different areas of Bangladesh. We have investigated several impacts of harassment on women in their life. In this regard, we have investigated below research question for our findings.

RQ1: At what age do women mostly face harassment?

RQ2: What are the main impacts they face when being harassed?

RQ3: What are the relation between age group and several impacts?

## VIII. Experimental Results and Discussions

We have used several python predefined libraries such as NumPy, pandas, matplotlib and apriori. CSV formatted data import to the system with the help of python predefined library pandas. At first, data was represented as pandas data frame format, and then it was converted in list format with the help of the NumPy library for making the data compatible with python. Then, the apriori and fp growth algorithms have been applied to the data set. In Fig.2, we organize all impacts based on the frequency of occurrences by using a minimum support value. Fig.3 shows the age frequency of respondents and represents the period when they mostly face harassment. Based on our research, it is clear that teenagers aged below 18 are most vulnerable to harassment. Based on our experiment, we have shown the relationship between impacts and respondent age. From Fig.4, it is clear that impacts mostly dominate over the teenager and respectively to other age groups. Fig.5 shows the graph of association between several harassment impacts and age groups. It is clear that several impacts made strong association rules among them and with age groups generated by association rule mining. From both association rules generated by the Apriori technique and FP-Growth technique, it is clear that the most vulnerable age group is teenagers that face harassment most. It is also shown that anxiety, depressions, intense fear, face difficulties in communication are the most frequent impacts that happen and generate strong association rules. According to association rules, it is also shown that many respondents face multiple impacts because of harassment. According to the results generated by the algorithm, it is shown that anxiety, depressions, ongoing fear have a strong association with age group under 18 individually, and again these impacts made association with other frequent impacts and with age group. In fig.6, the graph represents the association rules with minimum support and minimum confidence value, and with our actual value, some more random value is used to visualize the graph appropriately.

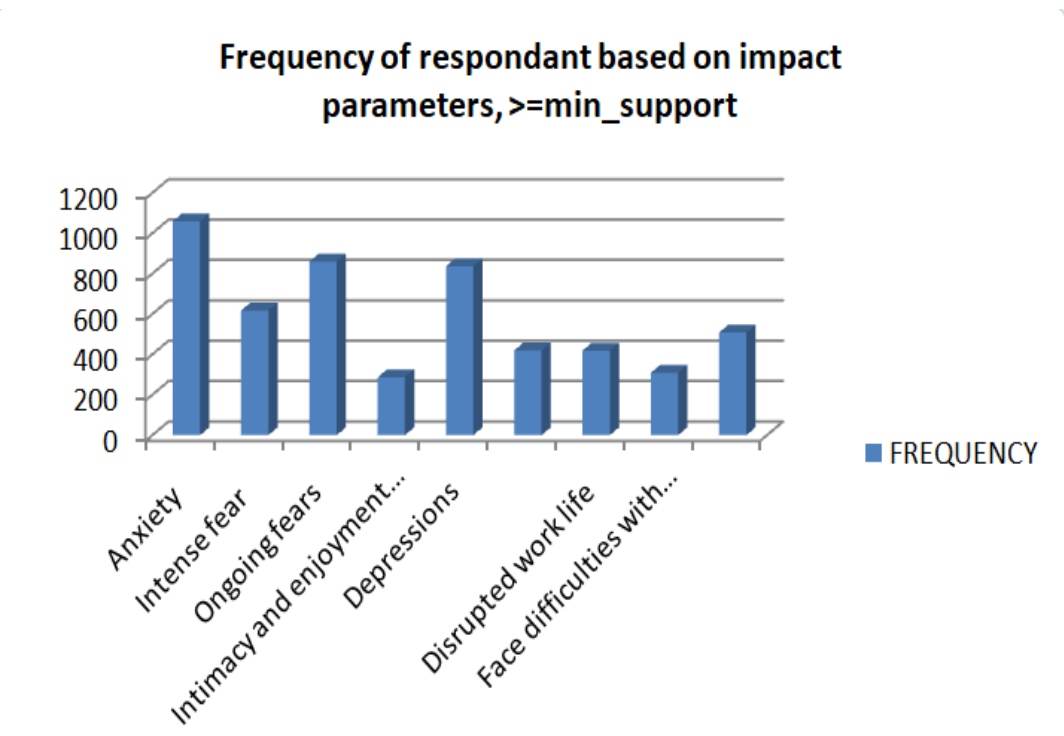

**Figure 2.**: All impacts based on the frequency of occurrences by using minimum support value.

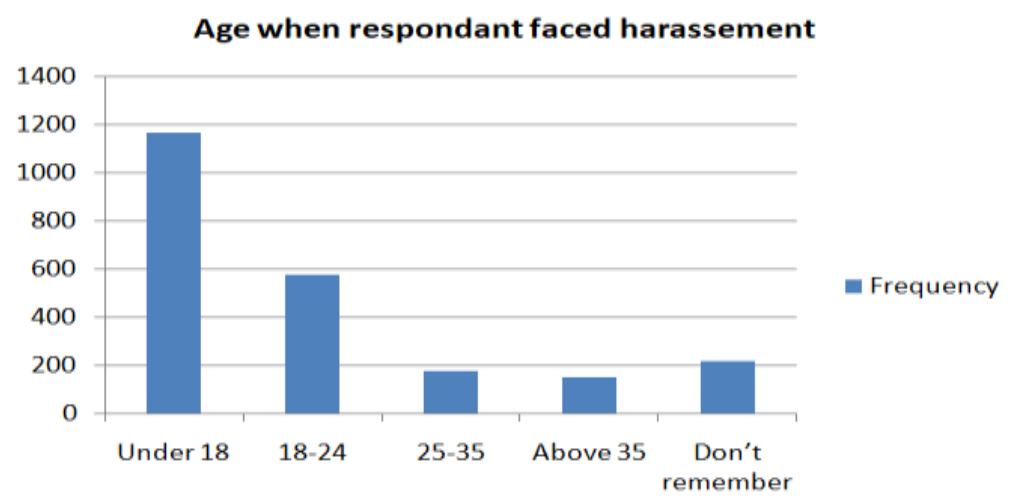

**Figure 3.** : Frequency of impacts by age.

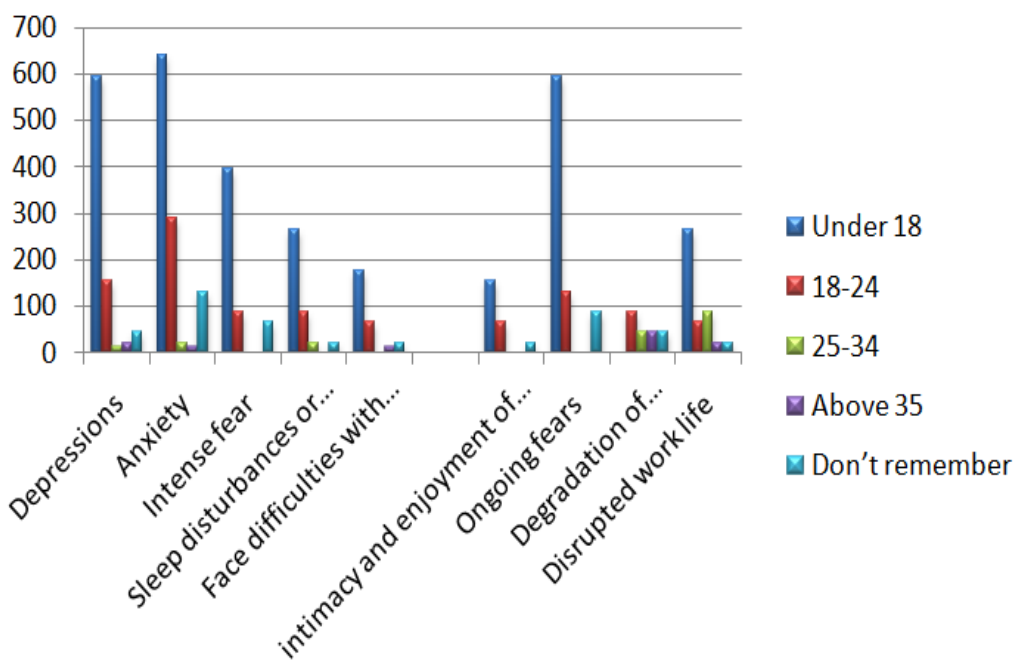

**Figure 4.** : Flow diagram of collected data

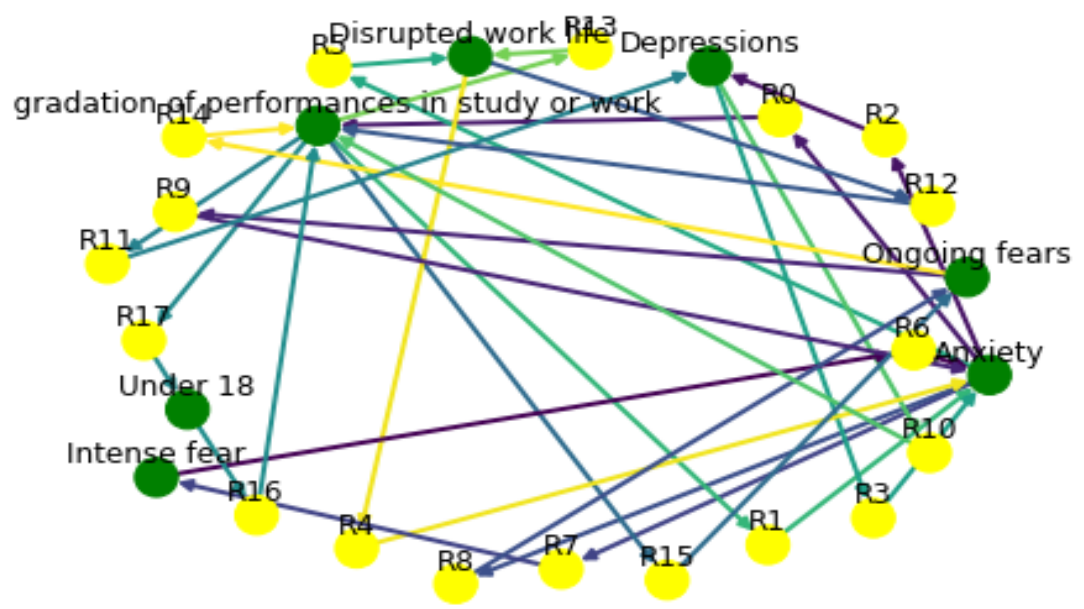

**Figure 5.:** Connectivity of association between attributes

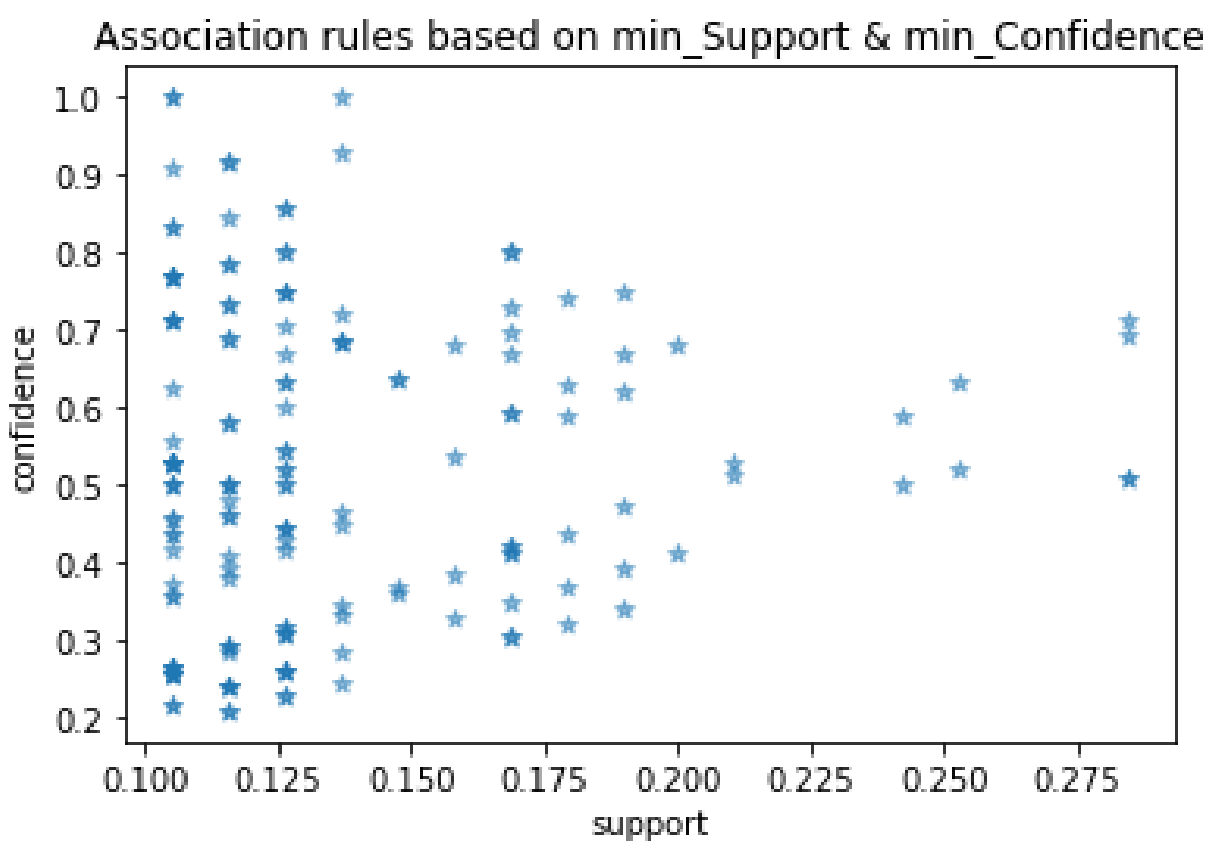

**Figure 6.** : Association based on min_sup. & confidence

## IX. Performance Analysis of Apriori and FP-Growth

Several parameters have different impacts on the performance of both algorithms. The parameters are broadly divided as minimum support value, the number of datasets, length of the individual dataset, and the number of individual data. Mostly F.P. Growth algorithm shows better performance than the Apriori algorithm. There are no candidate generation procedures in the F.P. Growth algorithm; rather, it creates a F.P. tree structure to store the items or data. When the number of datasets increased, the time taken for processing increased for both algorithms, although Apriori cost more time than F.P. Growth(Fig. 7). The length of individual datasets also impacts the performance of both algorithms. The processing time increased linearly with the increase of the dataset in the case of F.P. Growth but increased exponentially in the Apriori algorithm, which cost much more time in processing data. It is shown that the minimum support threshold value also affects the performance considerably in both algorithms (Fig. 8).

On the other hand, F.P. Growth takes more memory than Apriori because of its fp tree construction and construction of recursive conditional F.P. Tree. An increasing number of datasets increased the memory consumption in both algorithms. Length of datasets increased the consumption of memory in case of both algorithm and variation of data has less impact on memory. Additionally, the minimum support threshold caused less memory consumption.

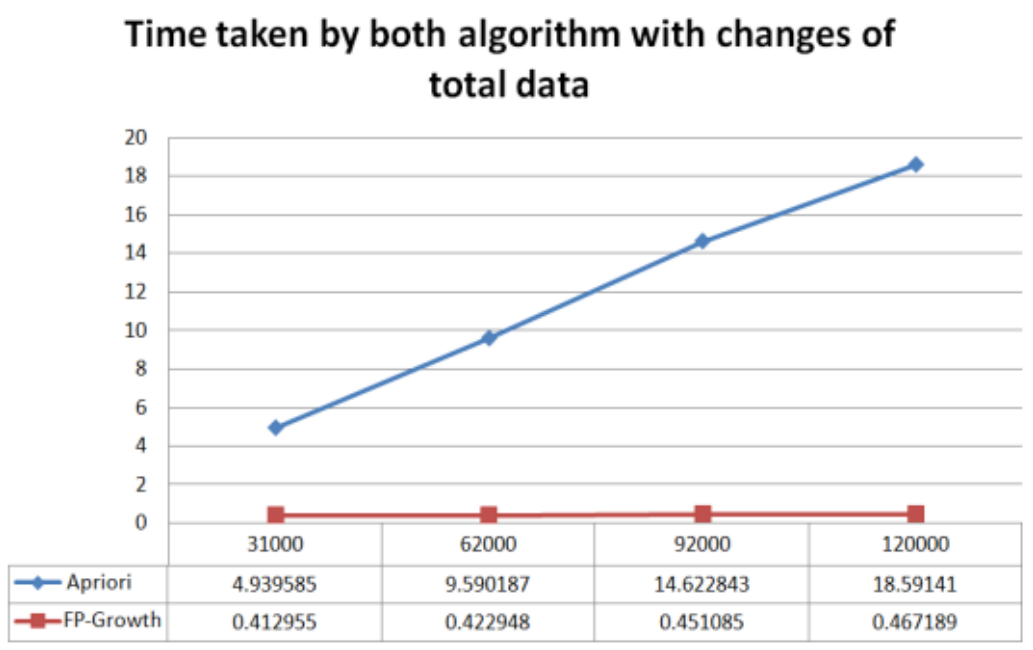

**Figure 7.**: Time performance based on number of attributes

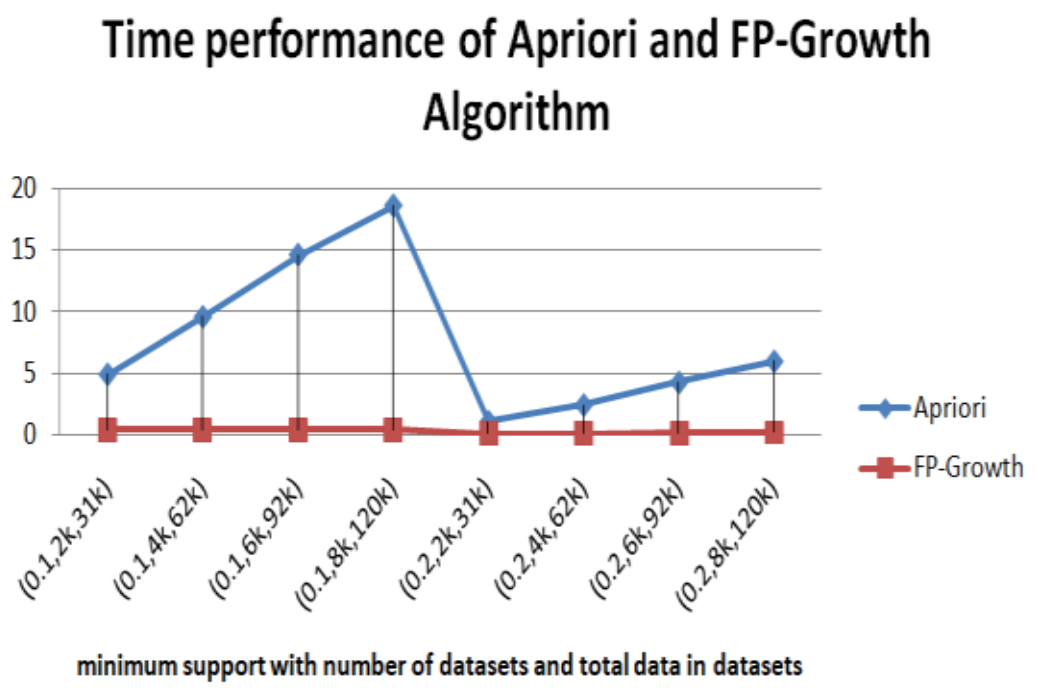

**Figure 8.** : Time performance based on minimum support

## X. Conclusions

It is clear that sexual harassment occurs in teenagers mostly. Most school and college-going girls are the main victims of harassment. They face harassment in public transport, street, in-home, and in their institutes also. Types of harassment and types of impacts differ on the basis of the age of the victim. It is also clear that impacts are severe when the victim is a teenager. However, Our research has considered limited data with limited respondents. Also, the coverage area is limited. On the other hand, the algorithms we used have several drawbacks to deal with. In the future, we will work with large volume data and other functionalities of association rule mining algorithms.